# Comparison of Patch-Based Conditional Generative Adversarial Neural Net Models with Emphasis on Model Robustness for Use in Head and Neck Cases for MR-Only Planning

## Authors:

**Peter Klages, Ilyes Bensilmane, Sadegh Riyahi, Jue Jiang, Margie Hunt, Joseph O. Deasy, Harini Veeraraghavan\*, Neelam Tyagi\***

**\*Co-Senior Authors**

**Acknowledgements:** This research was supported by Philips Healthcare under a Master Research Agreement and partially supported by the NIH/NCI Cancer Center Support Grant/Core Grant (P30 CA008748). Authors would like to acknowledge Dr. Reza Farjam for his assistance with his software and access to his previous data and results while he was at Memorial Sloan Kettering.

## Abstract

A total of twenty paired CT and MR images were used in this study to investigate two conditional generative adversarial networks, Pix2Pix, and Cycle GAN, for generating synthetic CT images for Head and Neck cancer cases. Ten of the patient cases were used for training and included such common artifacts as dental implants; the remaining ten testing cases were used for testing and included a larger range of image features commonly found in clinical head and neck cases. These features included strong metal artifacts from dental implants, one case with a metal implant, and one case with abnormal anatomy. The original CT images were deformably registered to the mDixon FFE MR images to minimize the processing of the MR images. The sCT generation accuracy and robustness were evaluated using Mean Absolute Error (MAE) based on the Hounsfield Units (HU) for three regions (whole-body, bone, and air within the body), Mean Error (ME) to observe systematic average offset errors in the sCT generation, and dosimetric evaluation of all clinically relevant structures. For the test set the MAEs for the Pix2Pix and Cycle GAN models were 92.4 ± 13.5 HU, and 100.7 ± 14.6 HU, respectively, for the whole-body region including tissue, bone, air, and artifacts, 166.3 ± 31.8 HU, and 184 ± 31.9 HU, respectively, for the bone region, and 183.7 ± 41.3 HU and 185.4 ± 37.9 HU for the air regions. The whole-body ME for the Pix2Pix and Cycle GAN models were 21.0 ± 11.8 HU and 37.5 ± 14.9 HU, respectively. Absolute

Percent Mean/Max Dose Errors were less than 2% for the PTV and all critical structures for both models, and DRRs generated from these models looked qualitatively similar to CT generated DRRs showing these methods are promising for MR-only planning.

# 1 Introduction

MR imaging provides superior soft-tissue contrast compared to CT imaging and is often considered the standard for tumor delineation in Head and Neck cancers (Dirix *et al.*, 2014). Its inclusion with CT imaging in radiotherapy (RT) dose planning has shown marked improvements in intra-observer tumor delineation, segmentation, and treatment outcomes in HN cancer (Dirix *et al.*, 2014; Rasch *et al.*, 2010; Chung *et al.*, 2004; Emami *et al.*, 2003). The current radiotherapy planning process requires delineation of target and organs-at-risk (OAR) structures on MR images followed by the transfer of these contours to CT images via image registration. This process introduces registration uncertainty. It is, therefore, desirable to move towards MR-only planning that could potentially reduce unnecessary CT imaging for patients, improve clinical workflow efficiency and reduce anatomical uncertainties stemming from registration errors between CT and MR images. However, the inability to directly calculate the electron density from MR images has, until recently, necessitated CT imaging for dose planning.

The methods and results for estimating Hounsfield Units (HU) and the electron density indirectly from MR images that have been developed over the last two decades are summarized in three recent review papers for different anatomical sites (Edmund and Nyholm, 2017; Johnstone *et al.*, 2018; Owrangi *et al.*, 2018). These review articles group the methods into voxel based, registration (atlas) based, and hybrid voxel-and-atlas-based classification. However, as computational techniques evolve, these categories become harder to fit uniquely, so it can also be helpful to categorize them by their core underlying processing methods. These categories are: manual segmentation, machine-learning, multi-atlas, and deep-learning based classification.

A central assumption of the methods to estimate HU values or electron density is that the distribution of MR intensities from the target and the existing set of training (or atlas) images are similar. This is essential for assigning HU intensities either using regression, as in a machine learning, or supervised clustering techniques, where clusters are assigned unique labels that have a corresponding CT intensity, as well as for the multi-atlas-based registration itself to work. However, MR images are not generally calibrated and typically depict variations from scan to scan, and scanner to scanner. Furthermore, the mapping from MR To CT tissue density is non-unique (not a one-to-one mapping) and, due to the short T2* relaxation time for cortical bone, bone is represented with similar image intensity values to air regions for almost all MR sequences. Ultrashort time echo (UTE) MR sequences can mitigate the ambiguity between air and bone since the ultrashort echo time permits bone measurements, but such acquisitions take considerably longer (>10 min) (Du *et al.*, 2014; Ma *et al.*, 2017) than pulse sequences such as mDixon (~5 min), thereby making UTE sequences less practical for clinical workflows. For non-UTE MR sequences, bias-field corrections, image standardization, machine learning and/or complex heuristics along with atlas techniques have proven effective in estimating structures. However, in addition to the accuracy of estimations, the practicality of these methods for a clinical workflow is also dependent on the associated manual and computational burden, as well as the robustness of the algorithms to previously unseen features.

With regards to the accuracy of the estimations, modern multi-atlas and deep learning methods are promising and active areas of research for synthetic CT (sCT) generation. In multi-atlas-based methods, the computational burden mainly lies in aligning the target MR scan with the set of MR images in the MR-CT atlas set. This process is computationally intense with times ranging from ten minutes to several hours to generate sCT images (Andreasen *et al.*, 2016; Arabi *et al.*, 2018; Farjam *et al.*, 2017; Torrado-Carvajal *et al.*, 2016; Uh *et al.*, 2014). Note also that computational time increases with the number of atlas pairs used. Ultimately, this computational burden makes these methods largely unsuitable for clinical practice at present, especially looking forward to where MR is used for online treatment adaptation. With deep learning methods, a convolutional neural network consisting of several convolutional layers is trained with data from the two modalities prior to sCT generation. While training is computationally intensive and may require several days, once trained, these methods are extremely fast and produce sCT in tens of seconds (Dinkla *et al.*, 2018; Emami *et al.*, 2018; Han, 2017; Nie *et al.*, 2018; Wolterink *et al.*, 2017). Another attractive feature of deep learning techniques is that they can directly extract the relevant set of features from the data without requiring extensive feature engineering.

Previous deep learning sCT generation methods have considered a variety of convolutional neural network models at largely two cancer sites: brain (Dinkla *et al.*, 2018; Emami *et al.*, 2018; Han, 2017; Nie *et al.*, 2018; Wolterink *et al.*, 2017) and pelvis (Chen *et al.*, 2018; Kim *et al.*, 2015; Maspero *et al.*, 2018; Nie *et al.*, 2017; Nie *et al.*, 2018; Savenije *et al.*, 2018). Of the methods reported, however, none of the prior deep learning studies, to the best of our knowledge, have focused on synthesizing sCT for the whole head and neck (HN) cancer cases. The most related studies instead only focus on the brain region, where deformable registration is less critical for aligning structures between multiple scans. We also note that prior sCT generation studies involving brain often excluded patient cases with strong streak artifacts in the CT images (a common occurrence for head and neck studies around mandible due to dental implants) (Dinkla *et al.*, 2018; Wolterink *et al.*, 2017). In this paper, we deliberately relaxed such exclusion criteria, and performed systematic evaluation on a variety of clinical conditions that included patients with significant dental artifacts, a case with fused vertebrae and a case with abnormal anatomy due to a very large tumor. We compared two conditional Generative Adversarial Network (GAN) based approaches: Pix2Pix (Isola *et al.*, 2017) and Cycle GAN (Zhu *et al.*, 2017). Both methods were trained using head and neck in-phase T1 fast field echo (FFE) mDixon MR images under the same conditions using image patches in three orthogonal views, enhanced training data augmentation, and three different post-processing techniques to integrate the overlapping patch-based inferences into a sCT. Finally, we evaluated the dosimetric consequences of generating sCT images with Cycle GAN and Pix2Pix neural networks, as well as generated digitally reconstructed radiographs to test the applicability of these neural networks for clinical usage.

## 1.1 Background: Generative Adversarial Networks (GANs)

GANs are a class of neural networks that pit two simpler convolutional neural networks, a generator and a discriminator, against each other to allow the generation of realistic looking images with similar statistical properties compared to a sample set from random noise, using only simple backpropagation and dropout in the training (Goodfellow *et al.*, 2014). The generator and discriminator are trained simultaneously using a min-max (minimize loss in generator and maximize discriminative capacity of discriminator) strategy that results in improved training stability for a wide range of content type, using the same network. These methods have been successfully applied to several applications ranging from synthesizing highly realistic examples for data augmentation in real-world images to synthetic CT generation in medical images.

Conditional GANs (Goodfellow *et al.*, 2014) extend the standard GAN approach such that instead of simply creating images with similar statistical properties, the synthesized images are derived with respect to the input images. Two predominant networks include the Pix2Pix paired image-to-image cGAN model (Isola *et al.*, 2017), and the cycle-consistent unpaired image-to-image model, Cycle GAN (Zhu *et al.*, 2017).

### 1.1.1 Image-to-Image Conditional Generative Adversarial Networks (Pix2Pix)

Pix2Pix (Isola *et al.*, 2017) is a conditional image-to-image generative adversarial network that requires paired CT and MR images from the same patient that are also co-registered with voxel-wise correspondence. The network consists of a generator and discriminator as used in the standard GAN model, but in addition to the standard GAN losses consisting of the generator loss (such as least square error or binary cross-entropy comparing the synthesized images with the training set) and the discriminator loss (real vs. fake), it includes an additional loss based on the absolute difference between the generated image and the original paired image (L1 norm loss). The generator network is a UNet (Ronneberger *et al.*, 2015) and the discriminator is composed of a sequence of convolutional layers.

### 1.1.2 Cycle-consistent Generative Adversarial Networks (Cycle GAN)

Cycle GAN (Zhu *et al.*, 2017) is related to Pix2Pix using the same basic network blocks, but it only needs images from each modality instead of near-perfectly aligned paired images from the same patient, as is required by Pix2Pix. This requirement is eliminated by imposing an additional cycle consistency loss that forces the network to produce images resembling those from the original modality (e.g. MR) using the synthesized images (e.g. sCT produced from MR) by minimizing the corresponding L1 norm losses between the synthesized and original modality images. A single cycle network is composed of a pair of GANs operating on the two imaging modalities. The generators are again UNets, and the discriminator networks are composed of five convolutional layers. Cycle consistency is required for both modality transformations. For example, with MR mDixon in-phase images and CT scans as the original modalities, the two cyclical networks are (MR→sCT$_1$→sMR$_1$) and (CT→sMR$_2$→sCT$_2$) with the aim that L1 losses for the MR and sMR$_1$ image pair and the CT and sCT$_2$ pair are minimized. This model is shown in Fig 1b.

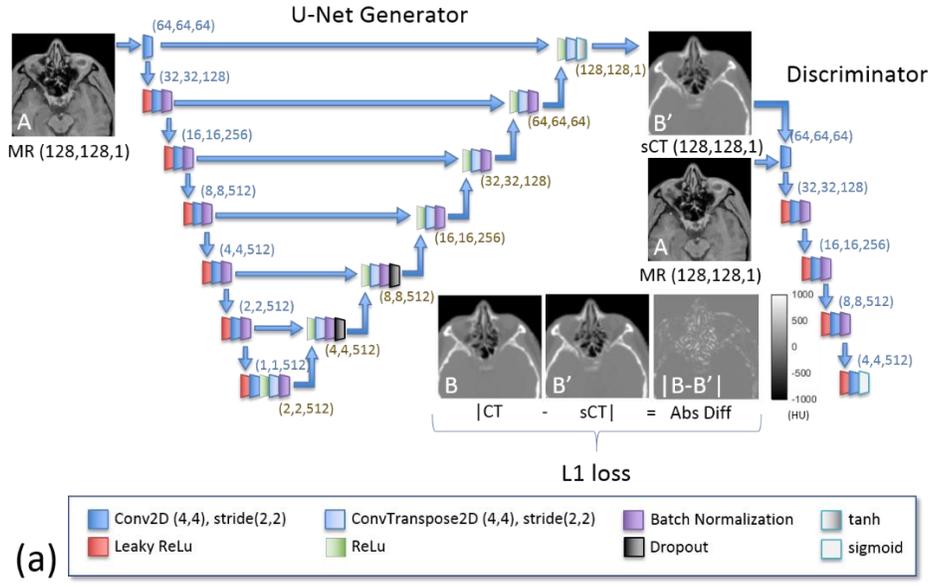

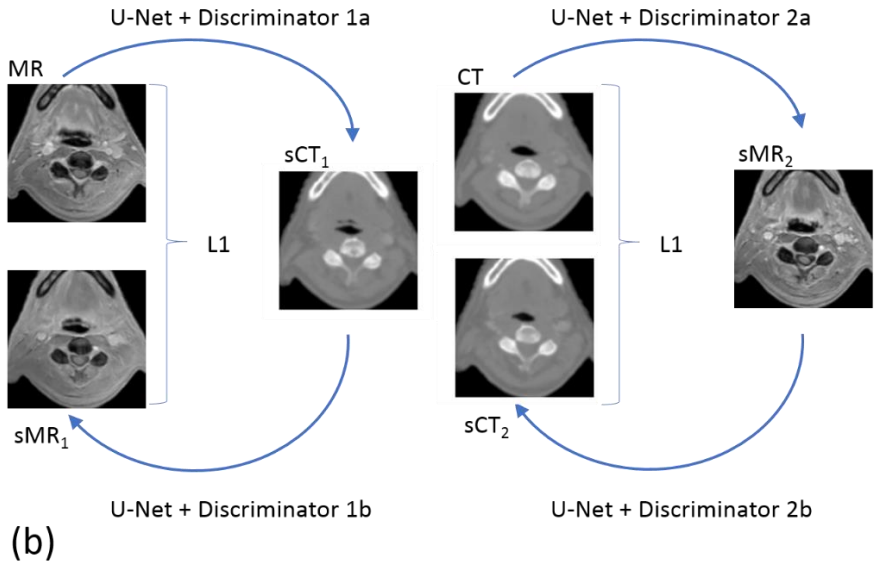

**Figure 1) (a)** The Pix2Pix model (Isola *et al.*, 2017) as used in this work, showing the generator U-Net, and the 5 level discriminator. In addition to these components, an L1 loss based on the synthetic and real CT is used. Paired images are required for Pix2Pix. **(b)** A simplified view of the Cycle GAN model (Zhu *et al.*, 2017) showing the two cycles comprised of Generators and Discriminators as from in (a) that are trained simultaneously to allow one to generate synthetic CT images without perfectly aligned CT and MR image pairs. An L1 loss between the input MR image and the synthetic MR image in the first loop, and an L1 loss between the CT and sCT in the second loop are used to further constrain the network.

## 2 Methods

In this study, we employed the Pix2Pix and Cycle GAN networks in a 2.5-D manner (three orthogonal views trained independently) with implementation enhancements on the data augmentation used for training, and a novel approach for producing HN sCT images for radiation treatment planning by combining overlapping sCTs patches. Our test cases included male and female patients, a wide range of ages, metal implants/artifacts, and unusual anatomy. We tested the sCT generation accuracy and robustness using Mean Absolute Error (MAE) based on Hounsfield Units (HU) for three regions (whole body, bone, and air within the body), use Mean Error (ME) to observe systematic average offset errors in the sCT generation, and evaluate the dosimetric uncertainties introduced by using sCTs. We also compared the accuracy of these deep learning methods to a multi-atlas sCT generation method (Farjam *et al.*, 2017).

## 2.1 Patient Selection

Twenty head and neck MR and CT image pairs from patients who received external beam radiotherapy were included in this study. There was no exclusion criterium aside from matching scan sequence parameters.

CT and MR image sets were acquired in radiotherapy treatment position. The MR images were acquired on a Philips 3T Ingenia system using the vendor-supplied phased-array dStream Head-Neck-Spine coil, with mDixon in-phase dual fast field echo (FFE) images (TE1/TE2/TR = 2.3/4.6/6.07 ms, flip angle = 10°, slice thickness of 2.4 mm and in-plane pixel size of approximately 1 mm$^2$). The CT images were acquired using a GE CT scanner in helical mode with tube voltage of 140 kV, and pitch factor 1.675.

A subset of ten patients were used to train the GAN networks and the remaining ten patients were used to test the trained networks. The cases for training were from the same set as used in our group's prior work on synthetic CT generation using multi-atlas methods (Farjam *et al.*, 2017). The testing set included two patients from our previous work (Farjam *et al.*, 2017) for continuity and direct comparison to those results, as well as eight additional cases not used in that study. These test cases included both high quality (minimal artifacts), to lower quality (patients with significant image artifacts including dental artifacts and a case with a metal implant) MR images to help test the deep learning network resiliency.

## 2.2 Image Processing

The workflow for processing images is shown in Figure 2. Image preprocessing steps are shown in Figure 2a and the workflow for the processed, registered CT/MR pairs for training and testing the two conditional GANs: Pix2Pix, and Cycle GAN, is shown in Figure 2b.

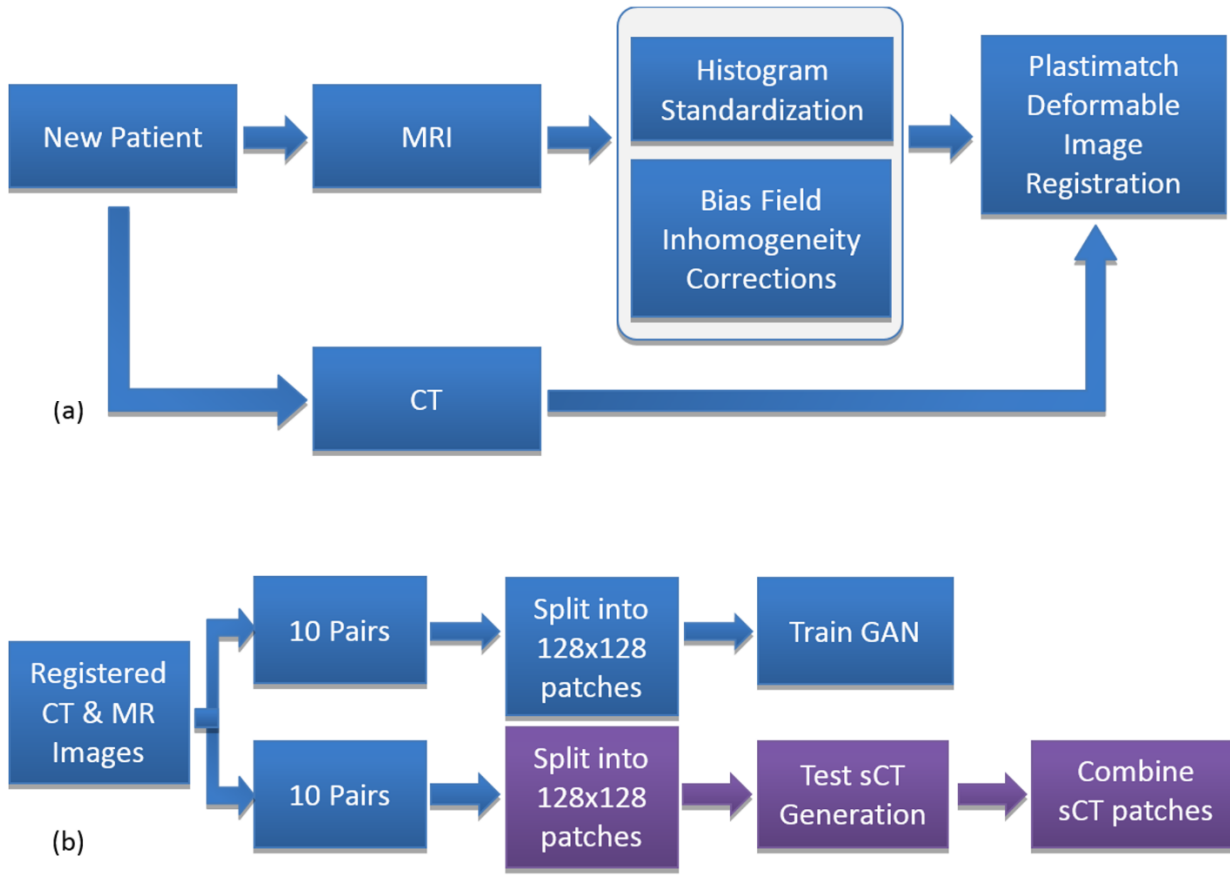

**Figure 2) (a)** Workflow for preprocessing the pairs of MR and CT images for training and testing: MR images have bias field corrections and histogram standardizations performed using vendor-provided software (CLEAR) and in-house software, then CT images are registered to the MR. **(b)** After preprocessing, the set of image pairs are divided into two groups of 10 pairs as described in the methods. The first set is used for training the neural networks and the second is for testing. Timing of the sCT generation is performed for the last three sections (marked purple).

### 2.2.1 MR Image Standardization and Bias Field Correction

The mDixon in-phase FFE MR image intensities may vary between scans and can also be affected by B0 and B1 field nonuniformities. We use vendor provided intensity inhomogeneity correction (CLEAR) and in-house developed software to reduce scanner-dependent intensity variations and bias field effects as described in our previous work (Farjam *et al.*, 2017).

### 2.2.2 Image Registration

The MR and CT images were both acquired in treatment position. The differences in patient setup position, head tilt/orientation, arm position (affecting the shoulder region of the scans) and different bite blocks were enough to preclude using only simple rigid registration. Multi-staged deformable image registration with Plastimatch (Sharp *et al.*, 2010) was used to register the CT images to the MR images for both the training and testing sets.

## 2.3 Training Details and Augmentation

The full volume scans (MR and CT) were resampled and padded with air values to produce isotropically spaced volume cubes of 512x512x512 voxels with voxel spacing of approximately 1 mm on edge, prior to any image scaling performed in the augmentation routines. Training of the neural networks was performed for all three views (axial, coronal, sagittal) individually. For the set of 10 cases used for training, the total number of unique, non-air only 512x512 pixel slices for the axial, coronal and sagittal views were 2663, 2524, and 4245 slices, respectively. To avoid training on large portions of air-filled regions, which could lead to trivial solutions, 128x128 pixel image patches were extracted from the 512x512 pixel image slices using a custom data extraction and augmentation function.

The custom data augmentation routines were performed on the image pairs at each epoch, where an epoch is defined as the computational period when all selected images patches have gone through the network optimization training and validation networks once. In addition to the exclusion of air portions, the augmentation is performed to help prevent overfitting in the network and to allow a greater variation of anatomical features and orientations which may be observed in new patient data. From each original $512 \times 512$ slice containing non-trivial data (i.e. slices containing more than just air), an additional mirror image along the symmetry axis of the axial and coronal views was created. Randomly chosen rotations in the range [-3.5, 3.5 degrees], scaling in the range [0.9,1.1], and shearing of the image in range [0.97, 1.03] were performed on each slice before 5 random cuts of $128 \times 128$ pixels were made. If no non-air pixels existed in a random cut, then it was discarded and up to 100 tries were performed before selecting an image pair patch. Thus, from an original set of 2663, 2524, and 4245 slices, for axial, coronal, and sagittal views, respectively, the network was trained on roughly $5.3 \times 10^6$, $5.0 \times 10^6$, and $8.5 \times 10^6$ image patches (128x128 voxels) after data augmentation over 200 epochs of training. Both the Pix2Pix and Cycle GAN models have over forty-one million parameters for the generator and over two million parameters for the discriminator in each view.

The Pix2Pix and Cycle GAN models were trained starting with the same original datasets, though they were augmented to different sets due to the random image modifications performed at each new epoch. The standardized, bias field corrected input images were clipped with a constant value of 2500 (>99.5 percentile intensity value of all masked voxels for the whole set of the standardized, BF corrected MR images) and scaled to the range [0,255] for input to the network and augmenter. The models use batch size of one for the input images, batch normalization, and the Adam optimizer (β1 = 0.5, β2 = 0.999) (Kingma and Ba, 2014) for the generator and discriminator networks.

## 2.4 Testing

The 512 × 512 × 512 voxel volumes with isotropic spacing (approximately 1 mm on edge) of the ten test cases were processed to create sets of 128 × 128 pixel images, with user specified strides that control the overlapping regions of the generated sCT estimation patches. Since accuracy toward the edges of the sCT patches was suspected to decrease from a lack of local context toward the outside of the image patch the number of border pixels to exclude when combining the output results was a tunable parameter specified by user as a pixel crop value. The parameters for stride, $s$, and cropping, $c$, are shown with the test workflow in Figure 3. We tested non-overlapping sCT image patches: stride 128 voxels, crop 0 voxels; and stride 96 voxels crop 16 voxels; as well as overlapping sCT regions: stride 32 voxels, crop 16 voxels; stride 32 voxels, crop 8 voxels.

Body masks for the MR images were created from the largest contiguous non-zero valued voxel regions. Holes in the body mask volumes were filled in automatically using a flood fill command for binary masks in Matlab, and the masks were dilated by 12 voxels (approximately 13.5 mm) to ensure that any boundary differences between the registered CT and the MR would not exclude portions of the sCT data by having too small a mask. Only image portions including masked regions were retained for processing to help speed up the sCT generation process.

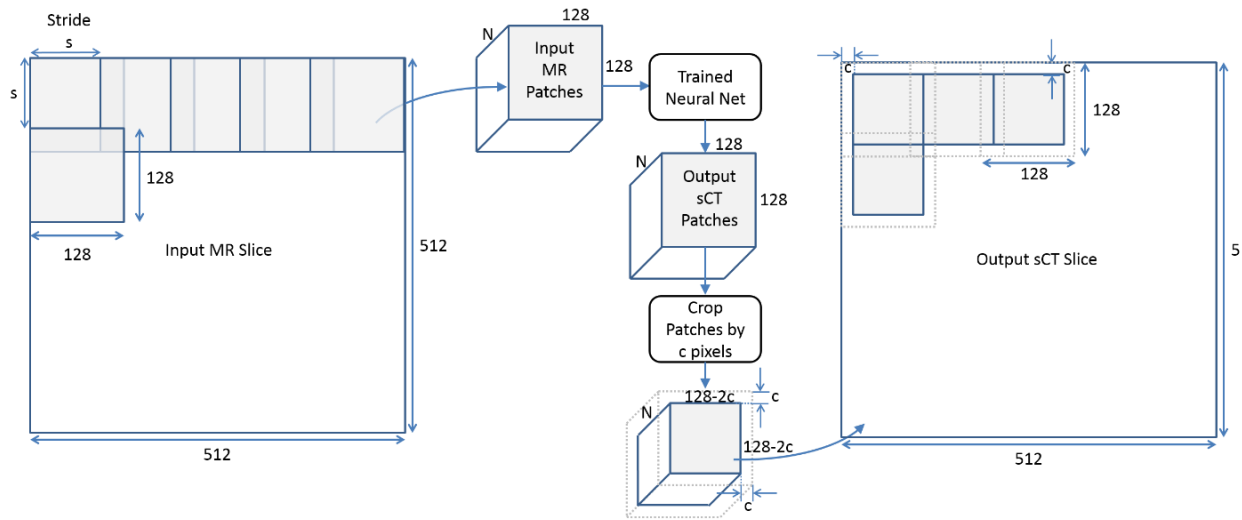

**Figure 3)** Image processing workflow showing the effects of changing the stride, $s$, on the number, N, of input patches and the cropped pixels on edge, $c$, when tiling the resultant sCT patches. This example case shows perfect tiling in the resultant sCT, but changes to s and/or c will result in overlapping sCT patches.

Three methods to combine the overlapping estimations of Hounsfield Units (HU) at a given voxel were investigated: averaging, median, and voting. Averaging and median work with the assumption of largely monomodal intensity distributions for a single voxel, while voting has the potential to perform better with multimodal HU intensity distributions. In voting, intensity values for a voxel are first classified into three types: air-like [-1000 HU, -200 HU), tissue-like [-200 HU, 200 HU), and bone/metal like [200 HU, 3071 HU]. After this classification of overlapping voxels into groups, the groups are voted on. Both majority-win and minority-win fraction criteria are defined prior to running the script. If the number of overlapping voxels in a classification group surpass the majority-win fraction (e.g. 65%) then the values from the other groups are discarded and the average of the values in the winning group is recorded for that voxel. If there is no majority winner, if the two largest groups surpass the minority fraction (e.g. 65% as well), then the values for the third group are discarded and the average of the values in the two winning groups are recorded for the voxel. We use 65% for both the majority and minority win fractions in this study.

### 2.5 Evaluation

Evaluation of the synthetic CT images can be broken into quantification of voxel-wise accuracy and overall dosimetric accuracy. The methods for evaluating the two aspects follow.

### 2.5.1 Mean Absolute Error, Mean Error, and Timing

The most common evaluation metrics for comparing new Multi-Atlas and convolutional neural network sCT generation techniques are the MAE and ME. Significant errors often occur in transition or interface regions with bone and air (such as the sinuses, internal ear structure, trachea, and esophagus). Thus, in addition to the MAE with standard deviation for the whole head and neck region, we also reported MAE for bone and air regions. These regions were chosen from the deformed CT images using HU thresholds of less than -300 HU [-1000 HU, -300 HU) and greater than 200 HU (200 HU, 3071 HU] for air-like and bone/metal regions, respectively. The body, air, and bone regions were all expanded by an integer number of voxels that yielded a distance less than 5 mm (4 voxels equals ~4.5 mm), to ensure that the whole transition zones were captured in the error metrics. Note that increasing the boundary zone too much would include soft tissue or air outside the body, which are relatively simple mappings in the networks, and would thus artificially lower the error for those zones.

Artifacts were not manually masked out in the evaluation of the sCT images compared to the CT images.

A detailed list of timestamps for the processing steps were recorded at testing.

### 2.5.2 Dosimetric Evaluation

To evaluate the accuracy of the HN sCTs for patient treatment planning, the treatment plan and structure set from the original planning CT was transferred to the sCT for all ten patients in the testing dataset and dose was recalculated and compared. Differences in patient setup between the planning CT and MR images prevented the direct transfer of structures between image sets. Therefore, the original planning structures were rigidly transferred to the deformed CT registered to the MR/sCT images using the previously described Plastimatch-based deformable registration method, and then when verified for the deformed CT, they were transferred to the sCT images. The following structures and dosimetric quantities were evaluated: PTV (max, D95), parotids (mean dose), submandibular glands (mean dose), brainstem (max dose), cord (max dose), and mandible (max dose).

Digitally reconstructed radiographs (DRRs) were also created to compare quality and accuracy of the bone structures for alignment purposes.

## 3 Results

### 3.1 Effects of Overlapping sCT Estimation Patches on MAE and ME

The MAE and ME for non-overlapped and overlapped sCT images using single view axial input images vs 2.5-D (three views: axial, sagittal, coronal) input MR images for both the Pix2Pix and Cycle GAN sCT generation methods are shown in Figure 4. MAE comparison volumes for the sCT and CT images are the body contours for the deformably registered CTs, expanded by four voxels, as described earlier in the methods. When using axial slices only to construct sCT images, the average MAE for the whole set of test patients was almost doubled when the naïve non-overlapping method was used (crop 0, stride 128), compared to the MAE from averaging the HU values from sCTs generated from the axial, sagittal, and coronal views. The MAE = $157.5 \pm 12.5$ HU vs $97.8 \pm 12.3$ HU for single vs multiple views, respectively, for the Pix2Pix model, and MAE = $167.2 \pm 13.3$ HU vs $107.0 \pm 13.0$ HU for single vs multiple views combined, respectively, for the Cycle GAN model. Similarly, the ME for the naïvely tiled axial-only sCT results are more than double the ME results for the combined three-view naïvely tiled results: $92.2 \pm 22.8$ HU vs $38.1 \pm 14.8$ HU for the Cycle GAN model and $74.6 \pm 22.1$ HU vs $23.3 \pm 11.2$ HU for the Pix2Pix model, respectively.

Cropping the sCT patches to reduce the number of voxel estimations that lack local spatial context (i.e. the patch border voxels) improves the MAE and ME for the resultant sCT images, but it is when sCT patches are overlapped and the voxel HU estimations are combined that the best sCTs are produced (Figure 4). When the resultant sCT patches are overlapped, the single-view axial results agree with the multi-view sCT estimations within one standard deviation for the MAE and ME results. Additionally, when using results comprised of sCT patches from the three views there is little difference between cropped multiple view non-overlapping sCT results (crop 16, stride 96) with the multiple view, overlapping results (they agree to within one standard deviation). However, the computational burden in using non-overlapping tiled sCT images for each view is greatly reduced compared to stacking, sorting and combining the relevant voxels in overlapping sCT patches (crop 8, stride 32 or crop 16, stride 32). With the overlapping method, there are as many as 16 estimations of a voxel HU value per view, for a total of as many as 48 estimations when all three views are combined for 2.5-D generation, versus only three estimations when they are tiled perfectly. Loading an input dataset, processing it, combining then saving the results for three, non-overlapping per-view sCT results takes approximately 1 min 10 s, with 51 s spent on setting up the networks and producing the sCT images for Pix2Pix, and 2 min 45 s with 2 min 30 s spent on setting up the networks and producing sCT images for Cycle GAN. As the number of sCT patches increases, the time to generate sCTs scales almost perfectly, but the sorting and combination functions for voting, averaging, or taking the median add additional processing and RAM. The non-optimized methods created for this project took an additional approximate 30 mins for the post-generation processing.

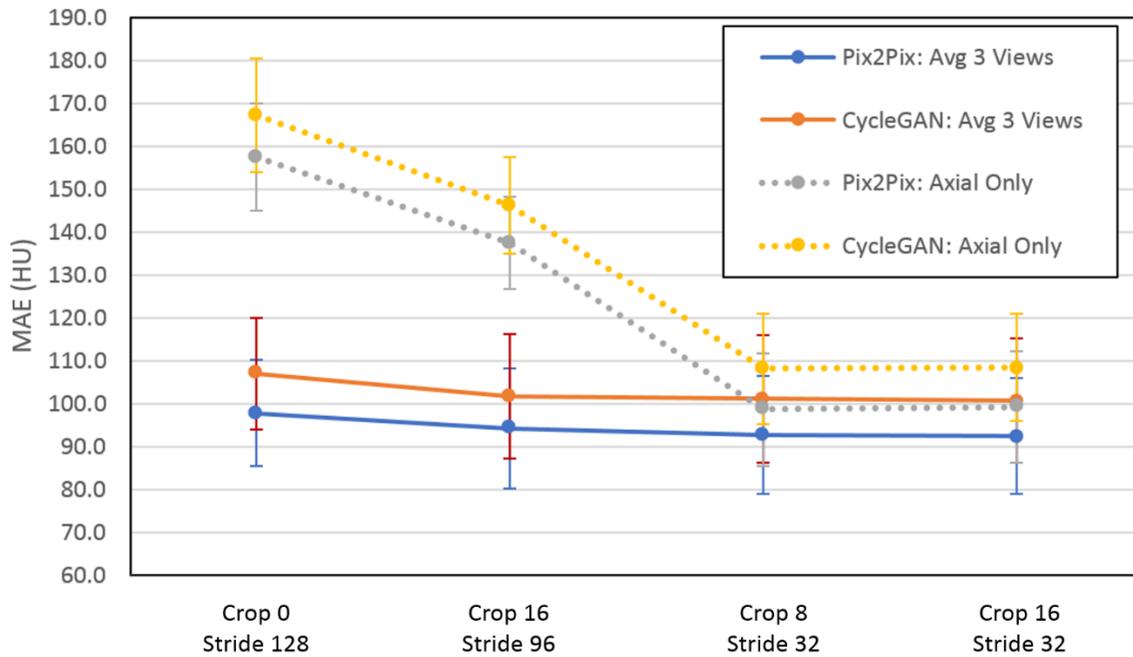

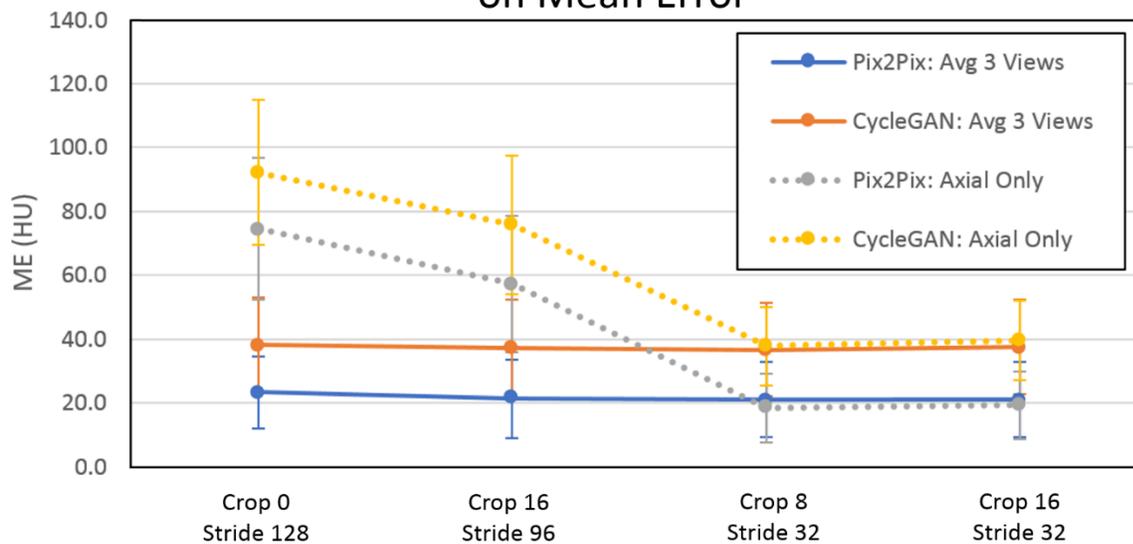

**Figure 4) (Top)** The Mean Absolute Error and **(Bottom)** Mean Error for the body region with respect to different crop and stride parameters, as described in the text. Both MAE and ME show the same general trends, namely that when multiple sCT estimations overlap and are averaged, either by using 3 complimentary views or overlapping patches in a single view, the errors decrease.

## 3.2 Effects of Intensity Standardization and Intensity Clipping on MAE and ME

The GANs were trained with MR intensities scaled within a range of [0, 255] which requires the test sets to also be scaled to the same range. The effects of MR histogram normalization at different MR image intensity clipping values on MAE and ME for the generated sCT images (using crop 16, stride 32 on the image patches) are presented in Table 1. The clip values are presented as three possibilities: dynamic, meaning the values are clipped at the 99-percentile intensity value for the masked voxels for each patient starting with the standardized, bias field corrected MR images; static: clipping at a predetermined intensity level of 2500; or dynamic, clipping at the 99$^{th}$ percentile intensity value for the masked values in the original uncorrected MR images. The results agree with each other within one standard deviation for each specific network model for both MAE and ME, showing a robustness against intensity and contrast variation.

**Table 1)** Effect of preprocessing input images on MAE and ME. The individual MR sets vary for 99 percentile intensity value, even after standardization, so changing from a dynamic to a static will affect the input image set contrast. This, and the relatively small changes in MAE and ME observed when using the original images, shows that the network is robust against contrast and intensity variation.

|  | Pix2Pix MAE (HU) | Std Dev (HU) | Cycle GAN MAE (HU) | Std Dev (HU) | Pix2Pix ME (HU) | Std Dev (HU) | Cycle GAN ME (HU) | Std Dev (HU) |
|---|---|---|---|---|---|---|---|---|
| Dynamic: 99 percentile on BFC, standardized MR sets | 96.6 | 11.5 | 104.4 | 14.0 | 22.0 | 13.0 | 38.9 | 16.2 |
| Static: 2500 clip value on BFC, standardized MR sets | 95.0 | 10.7 | 104.1 | 15.8 | 24.4 | 13.8 | 38.9 | 15.6 |
| Dynamic: 99 percentile on unmodified MR sets | 97.9 | 9.8 | 107.6 | 12.0 | 35.1 | 11.6 | 51.2 | 14.1 |

## 3.3 Effect of Merging Strategy Used to Combine Predictions from Neighboring Image Patches on MAE and ME

The MAE and ME for averaging, calculating the median, or voting (averaging the values from dominant classified ranges of values) to estimate the HU value for a voxel from overlapping patches (crop value = 16 pixels, stride = 32 pixels) with three orthogonal views (2.5-D sCT generation) agree within one standard deviation (Table 2). However, in the sCT images there are visible qualitative differences. The most notable differences are seen at the edges of bone and air sections, and in large uniform HU value regions: the edges of bones and air passages appear less blurred and the uniform regions appear less textured for the median and voting combination methods (Figure 5). Because the differences are subtle, and since averaging is the simplest, and thus quickest method to combine overlapping voxel estimations, the rest of the results are reported based on averaging method to combine the three-view sCT patches (crop value = 16 pixels, stride = 32 pixels).

**Table 2)** Effect of the combination method for overlapping voxel HU estimations for the 2.5-D, stride = 32, crop = 16 sCT generation. As many as 48 estimations per voxel are combined with the different methods.

| Combination Method | MAE | Std Dev | ME | Std Dev |
|---|---|---|---|---|
| Averaging of Values | 92.4 | 13.5 | 100.7 | 14.6 |
| Median of Values | 90.9 | 13.5 | 99.6 | 13.5 |
| Voting on Values | 91.5 | 13.5 | 100.4 | 13.5 |

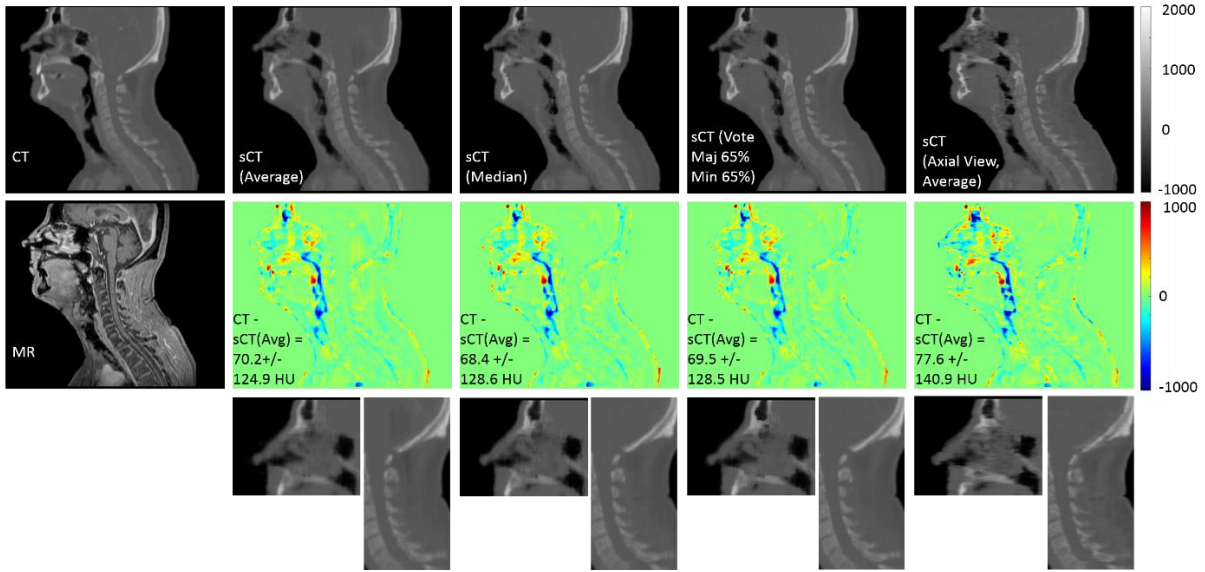

**Figure 5)** Comparison of effects of different combination methods (crop = 16, stride = 32). Averaging overlapping voxels blurs transition zone edges, calculating the median shows some noise at transitions as the estimations are not monomodally distributed, and voting (a combination of threshold-based classification and averaging the values for the majority classification) shows stronger edges than averaging. The sCT image, based off axial-view only patches, is also shown, illustrating how interplane estimations are not continuous by themselves (a sketchy texture is seen between layers). In the bottom row, zoomed in regions where edges and texture differences are most obvious are shown.

A summary of the results when averaging values to combine and using crop value = 16, stride = 32 voxels for the sCT patches for mean absolute error and mean error for the whole body, as well as MAE for the bone and air regions is shown in Table 3. The standard deviations for the individual masked volumes shows that there is a large range of values for MAE and ME for the ROIs, while the overall standard deviation for the set of patients is not very large.

Example sCT images generated using all three views, with the MR intensities clipped to the 99th percentile value following bias-field correction and histogram standardization and using the mean averaging from the overlapping image patches are shown for sagittal views in Figure 6 (top), and example results of the patient with a large tumor is shown in Figure 6 (bottom).

**Table 3)** Per-patient Mean Absolute Error (MAE) for whole head and neck region, bone region (threshold of 200 HU on original CT), and Air-like regions (< -200 HU) within the HN volume are 92.4 ± 13.5, 166.3 ± 31.8, and 183.7 ± 41.3 HU respectively for Pix2Pix, and 100.7 ± 14.6, 184.0 ± 31.9, and 185.4 ± 37.9 HU respectively for Cycle GAN. The Mean Error (the bias error) is 21.0 ± 11.8 and 37.5 ± 14.9 HU for Pix2Pix and Cycle GAN, respectively, showing that values from the real CT have higher valued HU values compared to the sCT image values, on average.

| | 3 View Combined | | | | Averaging mode | | | | Bones 200 HU threshold | | | | Air < -200 HU threshold | | | |
|---|---|---|---|---|---|---|---|---|---|---|---|---|---|---|---|---|
| | Body Abs Err | | | | Body Err | | | | Bones | | | | Air | | | |
| | Pix2Pix | | Cycle GAN | | Pix2Pix | | Cycle GAN | | Pix2Pix | | Cycle GAN | | Pix2Pix | | Cycle GAN | |
| Patient | MAE | std dev | MAE | std dev | ME | std dev | ME | std dev | MAE | std dev | MAE | std dev | MAE | std dev | MAE | std dev |
| 1 | 81.1 | 117.7 | 87.4 | 122.4 | 15.5 | 142.1 | 27.5 | 147.8 | 121.0 | 164.5 | 137.9 | 176.5 | 137.2 | 160.4 | 149.0 | 170.3 |
| 2 | 70.2 | 124.9 | 84.6 | 147.5 | 18.5 | 142.0 | 35.9 | 166.2 | 106.2 | 161.7 | 125.6 | 191.8 | 121.5 | 143.8 | 139.8 | 146.7 |
| 3 | 100.7 | 159.9 | 102.7 | 162.9 | 24.6 | 187.3 | 34.7 | 189.4 | 192.2 | 251.0 | 198.7 | 253.0 | 190.0 | 224.1 | 193.4 | 216.8 |
| 4 | 101.2 | 167.6 | 113.4 | 180.1 | 40.4 | 191.6 | 62.7 | 203.4 | 186.0 | 252.4 | 211.0 | 269.8 | 168.1 | 214.7 | 166.8 | 218.4 |
| 5 | 90.3 | 137.7 | 97.5 | 143.5 | 10.6 | 164.3 | 32.9 | 170.3 | 143.0 | 181.4 | 162.9 | 190.1 | 187.4 | 210.8 | 180.5 | 211.5 |
| 6 | 96.3 | 160.3 | 105.8 | 168.5 | 35.1 | 183.7 | 57.4 | 190.5 | 189.0 | 244.2 | 201.0 | 254.3 | 218.7 | 270.9 | 217.0 | 259.7 |
| 7 | 82.8 | 130.2 | 84.2 | 132.8 | 12.9 | 153.8 | 26.1 | 155.1 | 162.4 | 218.7 | 178.4 | 224.1 | 174.7 | 221.0 | 167.4 | 215.2 |
| 8 | 98.4 | 152.4 | 110.6 | 165.7 | 34.3 | 178.1 | 54.0 | 191.8 | 195.6 | 226.6 | 215.4 | 246.3 | 209.3 | 253.8 | 201.2 | 242.5 |
| 9 | 82.5 | 118.6 | 88.8 | 126.9 | 17.0 | 143.4 | 30.2 | 151.9 | 163.9 | 182.7 | 180.9 | 194.2 | 156.7 | 183.5 | 161.4 | 175.2 |
| 10 | 120.7 | 174.9 | 131.9 | 184.4 | 0.9 | 212.5 | 13.2 | 226.3 | 204.0 | 228.2 | 227.9 | 245.2 | 273.5 | 255.2 | 277.1 | 253.9 |
| avg | 92.4 | 144.4 | 100.7 | 153.5 | 21.0 | 169.9 | 37.5 | 179.3 | 166.3 | 211.1 | 184.0 | 224.5 | 183.7 | 213.8 | 185.4 | 211.0 |
| std dev | 13.5 | | 14.6 | | 11.8 | | 14.9 | | 31.8 | | 31.9 | | 41.3 | | 37.9 | |

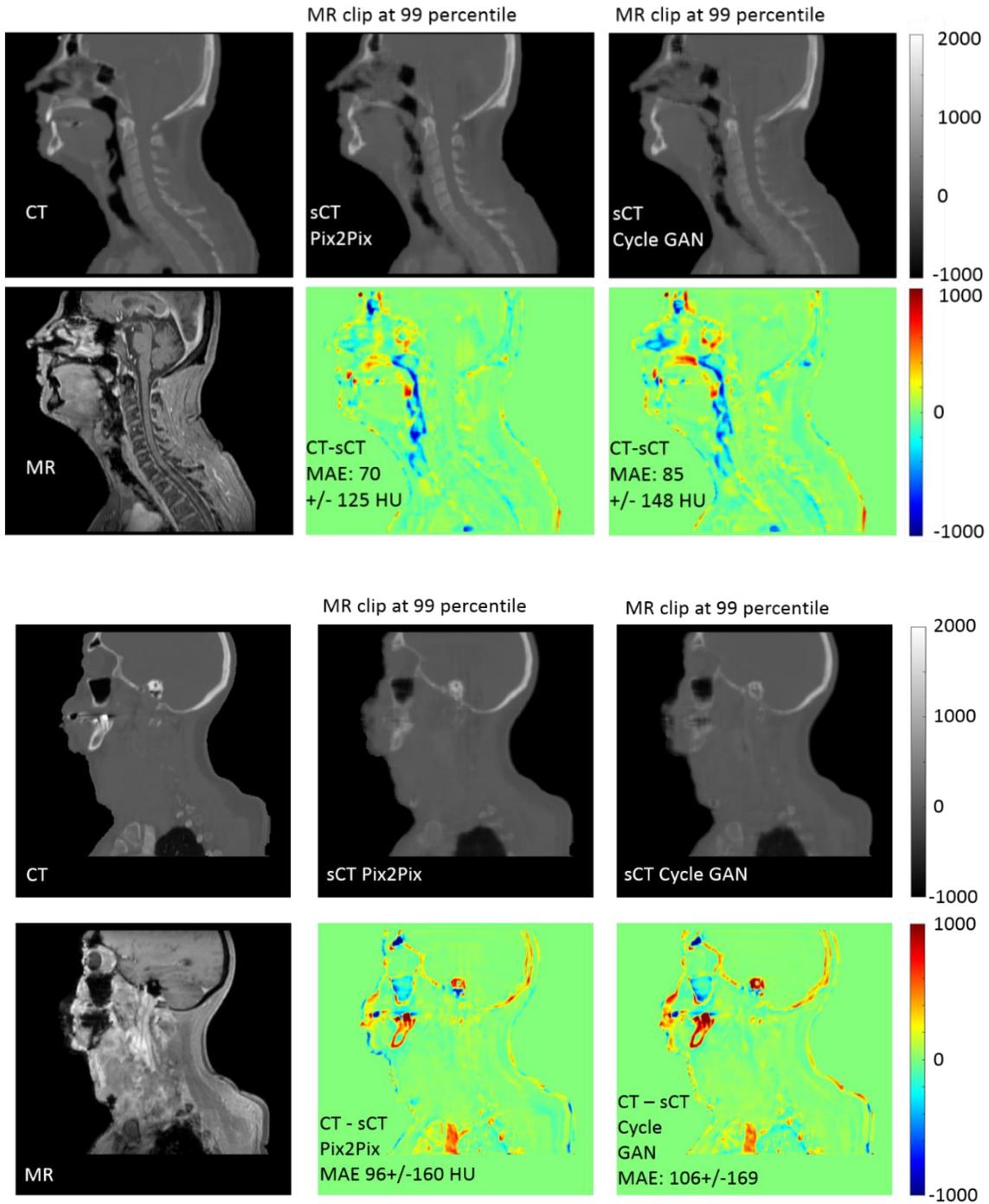

**Figure 6) (Top)** Example sCT images for case 2 with sCT patches from the three views combined by averaging, showing the similarities and differences between the resultant Pix2Pix and Cycle GAN sCT images. The most notable differences in the sagittal view are in the sinuses and nasopharynx, while in the axial view, the inner ear and sinuses show the most differences. **(Bottom)** Example sCT images for case 6, which included a large tumor. Both Pix2Pix and Cycle GAN estimate the HU values for the soft tissue of the tumor well, except for a small region at the anterior inferior part of the tumor recorded with low intensities in the MR image that was incorrectly transformed to bone-valued HUs.

## 3.4 Clinical and Dosimetric Results

An example dose volume histogram for Test Case 7 is shown in Figure 7. The sCTs for both the Pix2Pix and Cycle GAN networks agree to within 1% for all structures of interest in this case.

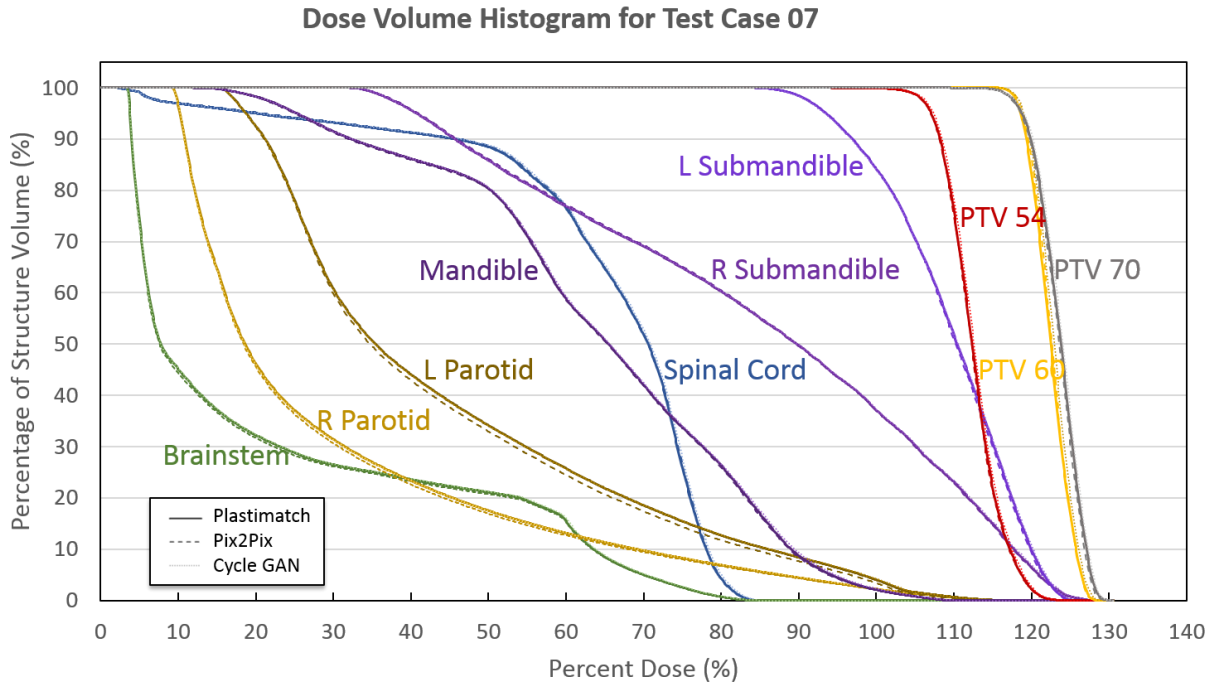

**Figure 7)** Dose Volume Histogram for all clinically relevant structures for test case 07, showing agreement to within 1% for the doses between the sCTs and the deformed CT. Original deformed CT DVHs are plotted with solid lines, Pix2Pix DVHs are plotted with dashed lines, and Cycle GAN DVHs are plotted with dotted lines.

The absolute dose difference and absolute percent dose difference between the Pix2Pix and Cycle GAN sCTs and the deformed CTs for all clinical structures of interest are presented in box whisker plots in Figure 8. Both the Pix2Pix and Cycle GAN models create clinically relevant sCT images, yielding absolute percent dose differences of 2% or less for all clinically relevant structures without discarding any outlier points, and less than 1.65% for all clinically relevant structures when one point in the mandible structure from a case with large dental artifacts is excluded (test case 10).

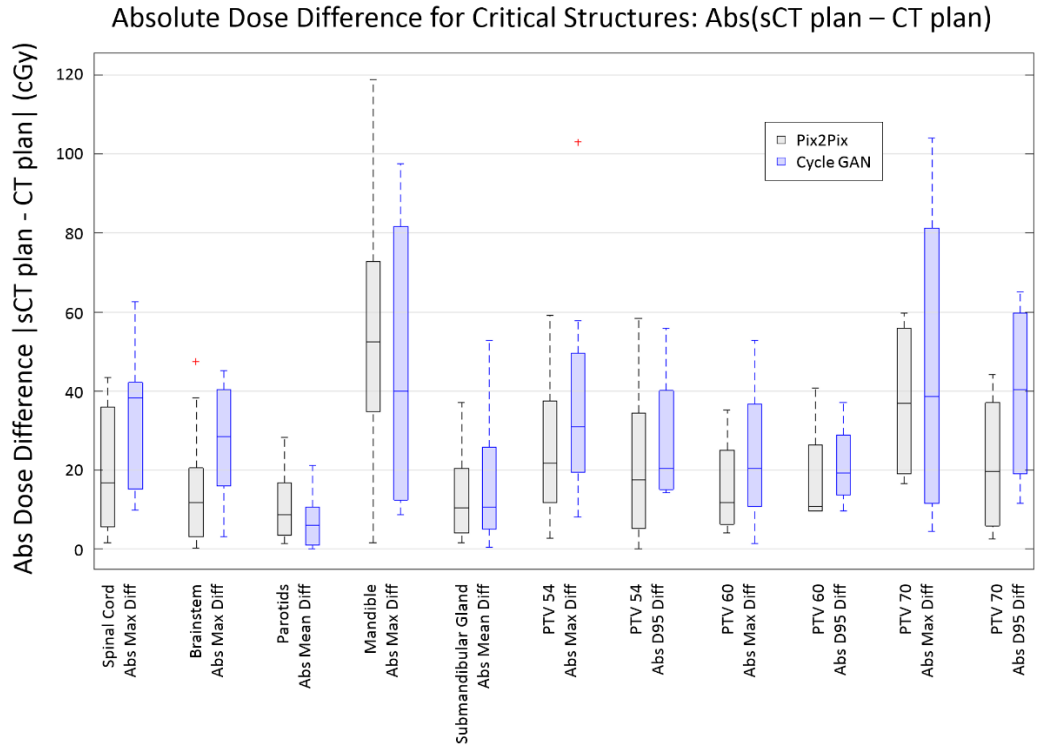

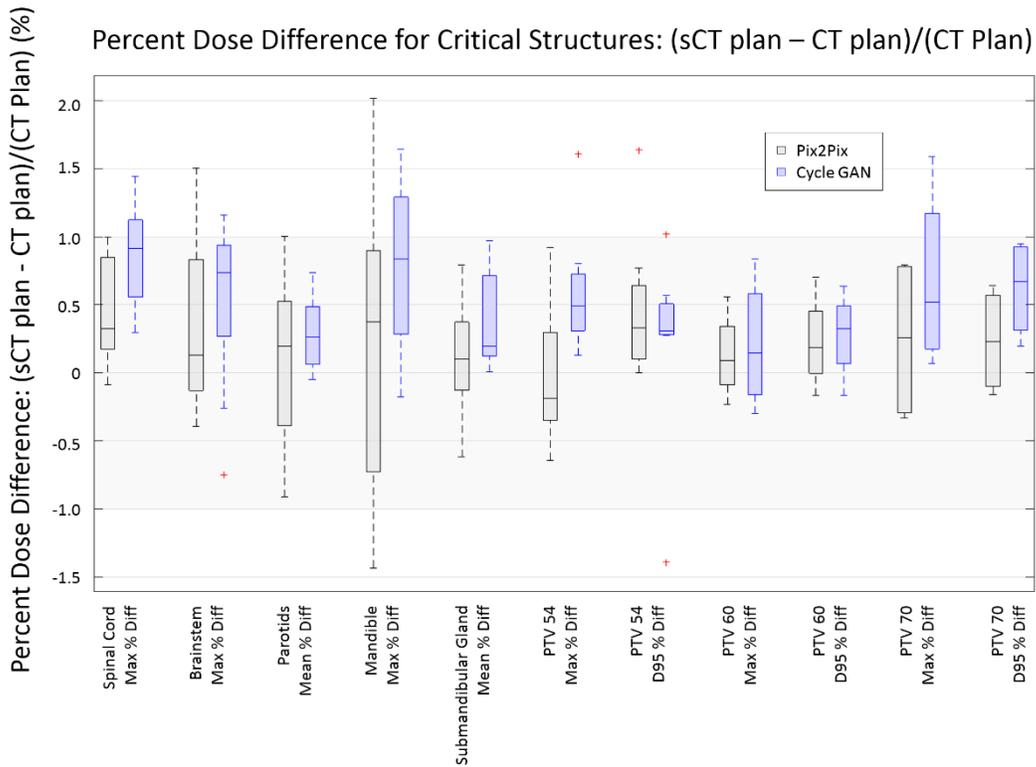

**Figure 8)** Box Whisker plots showing **(a)** Absolute Dose Difference and **(b)** Percent Dose Difference for all clinical structures of interest. The mandible, due to the dental artifacts, has the largest range of uncertainty of all clinically relevant structures for both Pix2Pix and Cycle GAN sCTs.

Example DRRs for test case 02 from the CT, Pix2Pix sCT, and Cycle GAN sCT images are shown in Figure 9. Differences between bone HU estimations and the HU values from the original CT lead to observable differences such as blurred edges and less intense bone structures, though the sCT images would still be useful for alignment purposes.

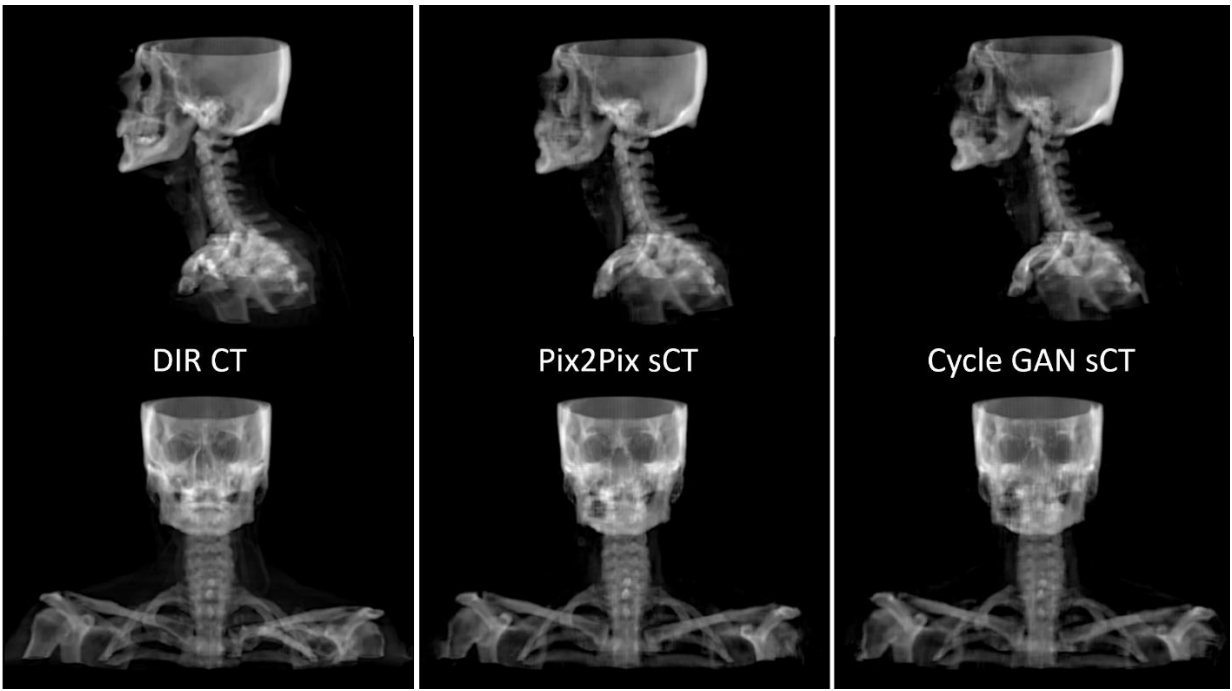

**Figure 9)** Sample sagittal and coronal view DRRs for the original CT, the Pix2Pix sCT, and the Cycle GAN sCT. Bone HU values are often underestimated compared to the original CT, and edges are blurred.

# 4 Discussion

Two conditional GANs were investigated for generating head and neck sCTs using the in-phase mDixon FFE sequence. The sCT generation methods were evaluated under conditions typically seen in our clinic including patients with significant dental artifacts leading to voids and local distortions in the MR images and streak artifacts on CTs, as well as different anatomy conditions including large tumors and body weights. Additionally, rigorous evaluation was performed under multiple image pre-processing (processing before analysis using GAN), post-processing (following sCT computation using GAN), as well as implementation conditions namely, one-view (axial) based sCT vs. multi-view (reformatted views consisting of axial, sagittal, and coronal) based sCT generation.

We found that training three separate neural networks, one for each orthogonal view to create a 2.5-D sCT generation method, resulted in lower MAE and ME and better constrained tissue extent compared with axial view training alone, even with in-plane overlapping of generated sCTs. We believe that the improvements in sCT accuracy are predominantly a result of the additional local spatial context provided by the three orthogonal views. Note that when using 2.5-D sCT generation, the difference between non-overlapping and multiple overlapping patches per-view is not significant based on the large standard deviations on the per-case MAE values, but the non-overlapping per-view method reduces the number of estimates by a factor of 16, resulting in reduced sCT generation times and simplified combination calculations. From loading the input dataset and initializing the neural network, to outputting the final processed dataset to disk, a total time of 1.5 minutes was needed for the 2.5-D non-overlapping per-view sCT generation.

With regards to the combination methods for overlapping sCT estimations at individual voxels, we tested three methods since it was not clear if the overlapping HU estimations for voxels would have monomodal or multimodal distributions. Averaging is the simplest combination method for overlapping voxel estimations, but the average value can be strongly affected by outlier HU values leading to variations in soft tissue and blurred transition regions between tissue types. Calculating the median value works well for suppressing outliers if the HU estimation distributions are monomodal, so soft tissue intensities are homogeneous, but at sinus and inner ear transitions between bone and air regions, the distribution may not be monomodal and the median value could lie between the two tissue types. Transition zones between tissue types are most strongly affected for the median calculation. Voting classifies voxels into general classes (air, soft-tissue, and bone/metal), and then calculates the average of the dominant tissue type(s). This suppresses noise in homogeneous soft-tissues and retains stronger edges at transition zones (Fig 5). The MAE and ME are not significantly different for the different combination methods, so despite the qualitative differences between the sCT images, the dosimetric results will be similar. Since the error results are so similar, we recommend the simplest, and thus fastest, combination method for general sCT creation with overlapping sCT patches.

The errors reported in this study were slightly higher compared to the most similar notable prior works (see Table 4) performed for brain sCT generation with deep learning. With no data sets to directly compare against, we speculate that the higher errors resulted in part from the different anatomical site (HN), and in part from our experimental conditions themselves, that included patients with observable artifacts on both CT and MR images and unusual anatomy. It is notable that the two cases in our test set that were identical to those reported in a prior study using multi-atlas registration by our group (Farjam *et al.*, 2017), had highly similar MAE ranges. Namely, when using the uniform intensity clipping value of 2500 for Pix2Pix sCT generation, the MAE was 68.4 HU and 65 HU for the two overlapping patients, respectively, and the overall

MAE value was 64 ± 10 HU reported in that study. These results suggest that the metal artifacts and unusual anatomy as considered in this study potentially contributed to our overall higher MAE of 92.4 ± 13.5 HU.

Our results show that despite using the same training images which were aligned the same way, compared to Pix2Pix, the Cycle GAN model still leads to worse MAE, ME, and dosimetric results, contrary to the results reported by Wolterink et al. (Wolterink *et al.*, 2017). However, there are notable implementation differences between our study and theirs, as they used single view (sagittal)-based sCT generation and rigid registration for aligning CT with MR images. While well-registered CT-MR pairs are not required for training Cycle GAN, they are still required to evaluate the results produced by these methods. Inconsistent or mis-registered images can adversely impact both training (in the case of Pix2Pix) and evaluation of error essential for training optimization and assessment of method performance, particularly when evaluating regions with bone and air interface. As pointed out in their study, mis-registrations between CT and MR images were present due to the rigid registrations, so we believe this is the reason why the Pix2Pix model results had poorer accuracy compared to the Cycle GAN model.

**Table 4)** Comparison of the studies with the most similar site (Brain) to our study (HN). The shaded cases had strong exclusion criteria, excluding cases with large streak/metal artifacts.

| Study | Year | NN Model | Location | MAE | Std Dev | ME | Std Dev | # Cases test/total |
|---|---|---|---|---|---|---|---|---|
| *Ours* | | *Pix2Pix* | *HN* | *92.4* | *13.5* | *21.0* | *11.8* | *10/20* |
| | | *Cycle GAN* | *HN* | *100.7* | *14.6* | *37.5* | *14.9* | *10/20* |
| (Dinkla *et al.*, 2018) | 2018 | Dilated CNN | Brain | 67 | 11 | 13 | 9 | 52 |
| (Emami *et al.*, 2018) | 2018 | ResNET | Brain | 89.3 | 10.3 | | | 15 |
| (Han, 2017) | 2017 | U-NET | Brain | 84.8 | 17.3 | -3.1 | 21.6 | 18 |
| (Nie *et al.*, 2018) | 2018 | 3D FCN GAN | Brain | 92.5 | 13.9 | | | 16 |
| (Wolterink *et al.*, 2017) | 2017 | Cycle GAN | Brain | 73.7 | 2.3 | | | 6/24 |
| | 2017 | Pix2Pix | Brain | 89.4 | 6.8 | | | 6/24 |

Nevertheless, both methods have clinically feasible dosimetric results for the cases tested, and the Cycle GAN method has the advantage of not requiring aligned CT-MR image pairs arising from the same patients which will simplify the preprocessing stages. This comes at the cost of increased training and sCT generation computational time, but if training sets are expected to grow or change over time when moving toward a clinical setting, this simplified preprocessing by relaxing the alignment criteria may be preferred.

One major advantage of both Pix2Pix and Cycle GAN methods, and deep learning methods in general, is that they learn the overall statistical characteristics of tissue correspondences between CT and MR images when producing the translation and are thus robust to presence of noise and small artifacts when generating sCT images. However, this modeling of overall statistical characteristics can also be a disadvantage as these methods do not necessarily preserve local spatial characteristics in the image, especially if training sets differ from test sets and if sufficiently strong loss criteria (such as L1 losses) are not used in training as shown in other works (Cohen *et al.*, 2018) including by our group (Jiang *et al.*, 2018). This defect/limitation was observable in an example patient case 6 with a very large tumor (see Figure 6, bottom), which had a small region incorrectly transformed to bone near the inferior part of the tumor growth by both methods since there were no example cases to transform the dark region of the tumor in the training stage.

Ultimately, the afore-mentioned limitation including mis-matching air cavities near bone regions could potentially be resolved with more inputs/different MR sequences including the UTE sequence, or higher-resolution MR images. Additionally, advanced methods such as in-painting techniques for easily identifiable artifacts may also lead to better results.

# 5 Conclusions

In this study, we evaluated two different state-of-the art conditional generative adversarial network models, namely, the Pix2Pix and Cycle GAN models, to generate synthetic CT images for head and neck cancer patients. Our results show that the Pix2Pix method slightly outperformed the Cycle GAN method. Furthermore, we found that overlapping HU estimations from patches with different spatial contexts strengthens the overall sCT generation, especially integrating multiple imaging views for training (2.5-D) vs. single views. The neural network models were shown to be robust against input MR image intensities, small variations in contrast, and were also shown to work with non-standardized, non-bias-field-corrected input MR images. By learning statistical characteristics of the source and target modalities, these models can generate sCTs for previously untrained features, such as large or unusual tumors, but we urge caution in evaluating cases with unique features or non-standard MR artifacts.

# References

Andreasen D, Van Leemput K and Edmund J M 2016 A patch-based pseudo-CT approach for MRI-only radiotherapy in the pelvis *Med Phys* **43** 4742-52

Arabi H, Dowling J A, Burgos N, Han X, Greer P B, Koutsouvelis N and Zaidi H 2018 Comparative study of algorithms for synthetic CT generation from MRI: Consequences for MRI-guided radiation planning in the pelvic region *Med Phys* **45** 5218-33

Chen S, Qin A, Zhou D and Yan D 2018 Technical Note: U-net-generated synthetic CT images for magnetic resonance imaging-only prostate intensity-modulated radiation therapy treatment planning *Med Phys* **45** 5659-65

Chung N-N, Ting L-L, Hsu W-C, Lui L T and Wang P-M 2004 Impact of magnetic resonance imaging versus CT on nasopharyngeal carcinoma: primary tumor target delineation for radiotherapy *Head & Neck* **26** 241-6

Cohen J P, Luck M and Honari S 2018 Distribution Matching Losses Can Hallucinate Features in Medical Image Translation. pp 529-36

Dinkla A M, Wolterink J M, Maspero M, Savenije M H F, Verhoeff J J C, Seravalli E, Isgum I, Seevinck P R and van den Berg C A T 2018 MR-Only Brain Radiation Therapy: Dosimetric Evaluation of Synthetic CTs Generated by a Dilated Convolutional Neural Network *Int J Radiat Oncol* **102** 801-12

Dirix P, Haustermans K and Vandecaveye V 2014 The Value of Magnetic Resonance Imaging for Radiotherapy Planning *Seminars in Radiation Oncology* **24** 151-9

Du J, Ma G, Li S, Carl M, Szeverenyi N M, VandenBerg S, Corey-Bloom J and Bydder G M 2014 Ultrashort echo time (UTE) magnetic resonance imaging of the short T2 components in white matter of the brain using a clinical 3T scanner *NeuroImage* **87** 32-41

Edmund J M and Nyholm T 2017 A review of substitute CT generation for MRI-only radiation therapy *Radiation Oncology* **12** 28

Emami B, Sethi A and Petruzzelli G J 2003 Influence of MRI on target volume delineation and IMRT planning in nasopharyngeal carcinoma *International Journal of Radiation Oncology, Biology, Physics* **57** 481-8

Emami H, Dong M, Nejad-Davarani S P and Glide-Hurst C K 2018 Generating synthetic CTs from magnetic resonance images using generative adversarial networks *Med Phys* **45** 3627-36

Farjam R, Tyagi N, Veeraraghavan H, Apte A, Zakian K, Hunt M A and Deasy J O 2017 Multiatlas approach with local registration goodness weighting for MRI-based electron density mapping of head and neck anatomy *Med Phys* **44** 3706-17

Goodfellow I, Pouget-Abadie J, Mirza M, Xu B, Warde-Farley D, Ozair S, Courville A and Bengio Y 2014 Generative Adversarial Nets 2672--80

Han X 2017 MR-based synthetic CT generation using a deep convolutional neural network method *Med Phys* **44** 1408-19

Isola P, Zhu J-Y, Zhou T and Efros A A *2017 IEEE Conference on Computer Vision and Pattern Recognition (CVPR),2017),* vol. Series*)*: IEEE) pp 5967-76

Jiang J, Hu Y-C, Tyagi N, Zhang P, Rimner A, Mageras G S, Deasy J O and Veeraraghavan H *International Conference on Medical Image Computing and Computer-Assisted Intervention,2018),* vol. Series*)*: Springer) pp 777-85

Johnstone E, Wyatt J J, Henry A M, Short S C, Sebag-Montefiore D, Murray L, Kelly C G, McCallum H M and Speight R 2018 Systematic Review of Synthetic Computed Tomography Generation Methodologies for Use in Magnetic Resonance Imaging–Only Radiation Therapy *International Journal of Radiation Oncology, Biology, Physics* **100** 199-217

Kim J, Glide-Hurst C, Doemer A, Wen N, Movsas B and Chetty I J 2015 Implementation of a Novel Algorithm For Generating Synthetic CT Images From Magnetic Resonance Imaging Data Sets for Prostate Cancer Radiation Therapy *International Journal of Radiation Oncology, Biology, Physics* **91** 39-47

Kingma D P and Ba J 2014 Adam: A Method for Stochastic Optimization *CoRR* **abs/1412.6980**

Ma Y-J, Carl M, Shao H, Tadros A S, Chang E Y and Du J 2017 Three-dimensional ultrashort echo time cones T1ρ (3D UTE-cones-T1ρ) imaging *NMR in Biomedicine* **30** e3709

Maspero M, Savenije M H F, Dinkla A M, Seevinck P R, Intven M P W, Jurgenliemk-Schulz I M, Kerkmeijer L G W and van den Berg C A T 2018 Dose evaluation of fast synthetic-CT generation using a generative adversarial network for general pelvis MR-only radiotherapy *Phys Med Biol* **63**

Nie D, Trullo R, Lian J, Petitjean C, Ruan S, Wang Q and Shen D *Medical Image Computing and Computer-Assisted Intervention – MICCAI 2017, (Cham, 2017// 2017),* vol. Series) ed M Descoteaux*, et al.*: Springer International Publishing) pp 417-25

Nie D, Trullo R, Lian J, Wang L, Petitjean C, Ruan S, Wang Q and Shen D 2018 Medical Image Synthesis with Deep Convolutional Adversarial Networks *IEEE Transactions on Biomedical Engineering* 1-

Owrangi A M, Greer P B and Glide-Hurst C K 2018 MRI-only treatment planning: benefits and challenges *Physics in Medicine & Biology* **63** 05TR1

Rasch C R, Steenbakkers R J, Fitton I, Duppen J C, Nowak P J, Pameijer F A, Eisbruch A, Kaanders J H, Paulsen F and van Herk M 2010 Decreased 3D observer variation with matched CT-MRI, for target delineation in Nasopharynx cancer *Radiation Oncology* **5** 21

Ronneberger O, Fischer P and Brox T 2015 U-Net: Convolutional Networks for Biomedical Image Segmentation *Lect Notes Comput Sc* **9351** 234-41

Savenije M H F, Maspero M, Dinkla A M, Seevinck P R and Van den Berg C A T 2018 MR-based synthetic CT with conditional Generative Adversarial Network for prostate RT planning *Radiother Oncol* **127** S151-S2

Sharp G C, Li R, Wolfgang J, Chen G, Peroni M, Spadea M F, Mori S, Zhang J, Shackleford J and Kandasamy N *Proceedings of the XVI'th International Conference on the use of Computers in Radiotherapy (ICCR), Amsterdam, Netherlands,2010),* vol. Series)

Torrado-Carvajal A, Herraiz J L, Alcain E, Montemayor A S, Garcia-Canamaque L, Hernandez-Tamames J A, Rozenholc Y and Malpica N 2016 Fast patch-based pseudo-CT synthesis from T1-weighted MR images for PET/MR attenuation correction in brain studies *Journal of Nuclear Medicine* **57** 136-43

Uh J, Merchant T E, Li Y, Li X and Hua C 2014 MRI-based treatment planning with pseudo CT generated through atlas registration *Med Phys* **41** 051711

Wolterink J, Dinkla A, H. F. Savenije M, Seevinck P, Berg C and Išgum I 2017 *Deep MR to CT Synthesis Using Unpaired Data*

Zhu J, Park T, Isola P and Efros A A *Ieee I Conf Comp Vis,22-29 Oct. 2017 2017),* vol. Series) pp 2242-51